\documentclass[journal]{IEEEtran}
\usepackage{tabularx}
\usepackage{multirow}
\usepackage{cite}
\usepackage{graphicx} 
\usepackage{multirow}  
\usepackage{amsmath}
\usepackage{algorithm}
\usepackage{algorithmic}
\usepackage{xcolor}
\begin{document}

\title{How Good is ChatGPT at Audiovisual Deepfake Detection: A Comparative Study of ChatGPT, AI Models and Human Perception}

\author{Sahibzada Adil Shahzad, Ammarah Hashmi, Yan-Tsung Peng, Yu Tsao,~\IEEEmembership{Senior Member,~IEEE}, Hsin-Min Wang,~\IEEEmembership{Senior Member,~IEEE}
\thanks{Sahibzada Adil Shahzad is with the Social Networks and Human-Centered Computing Program, Taiwan International Graduate Program, Academia Sinica, Taipei, Taiwan, and also with the Department of Computer Science, National Chengchi University, Taipei, Taiwan. (e-mail: adilshah275@iis.sinica.edu.tw).

Ammarah Hashmi is with the Social Networks and Human-Centered Computing Program, Taiwan International Graduate Program, Academia Sinica, Taipei, Taiwan, and also with the Institute of Information Systems and Applications, National Tsing Hua University, Hsinchu, Taiwan. (e-mail: hashmiammarah0@gmail.com).

Yan-Tsung Peng is with the Department of Computer Science, National Chengchi University, Taipei, Taiwan. (e-mail: ytpeng@cs.nccu.edu.tw)

Yu Tsao is with the Research Center for Information Technology Innovation, Academia Sinica, Taipei, Taiwan. (e-mail:
yu.tsao@citi.sinica.edu.tw).

Hsin-Min Wang is with the Institute of Information Science, Academia Sinica, Taiwan. (e-mail: whm@iis.sinica.edu.tw)

}}


\maketitle

\begin{abstract}
Multimodal deepfakes involving audiovisual manipulations are a growing threat because they are difficult to detect with the naked eye or using unimodal deep learning-based forgery detection methods. Audiovisual forensic models, while more capable than unimodal models, require large training datasets and are computationally expensive for training and inference. Furthermore, these models lack interpretability and often do not generalize well to unseen manipulations. In this study, we examine the detection capabilities of a large language model (LLM) (i.e., ChatGPT) to identify and account for any possible visual and auditory artifacts and manipulations in audiovisual deepfake content. Extensive experiments are conducted on videos from a benchmark multimodal deepfake dataset to evaluate the detection performance of ChatGPT and compare it with the detection capabilities of state-of-the-art multimodal forensic models and humans. Experimental results demonstrate the importance of domain knowledge and prompt engineering for video forgery detection tasks using LLMs. Unlike approaches based on end-to-end learning, ChatGPT can account for spatial and spatiotemporal artifacts and inconsistencies that may exist within or across modalities. Additionally, we discuss the limitations of ChatGPT for multimedia forensic tasks.
\end{abstract}


\section{Introduction}
\begin{figure}[!ht]
\centering
\fbox{\includegraphics[width=0.47\textwidth]{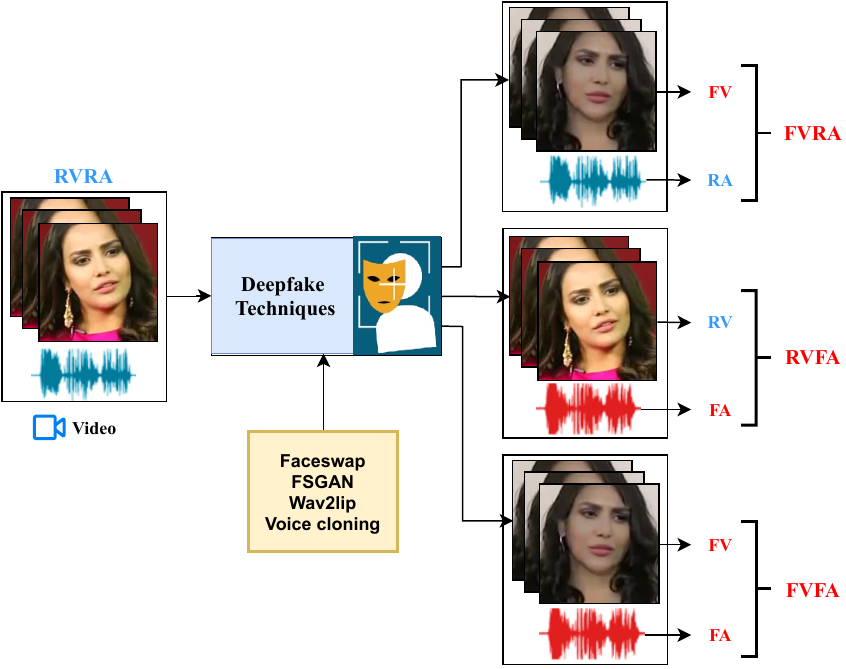}}
\caption{Illustration of audiovisual deepfake manipulations. Original video content is represented as RVRA (real video with real audio. Through deepfake manipulation techniques, three manipulated types are generated: FVRA (fake video with real audio), RVFA (real video with fake audio), and FVFA (fake video with fake audio). Blue text represents the ``real modality'' of the video content, while red text represents the ``fake modality''.}
\label{fig:Fig1}
\vspace{-4mm}
\end{figure}

Synthetic multimedia content has become both innovative and a significant threat in recent years. Deepfake images and videos created using artificial intelligence (AI) and deep learning (DL) techniques have attracted public and academic attention. This synthetic content is generated by generative adversarial networks (GANs)~\cite{goodfellow2020generative} and more sophisticated AI techniques such as diffusion models~\cite{ho2020denoising}. While deepfake technology has many innovative applications in education, entertainment, and other fields~\cite{kwok2021deepfake}, it is a double-edged sword that can be used for unethical purposes, such as pornography, political defamation, identity theft, fraud, misinformation, and disinformation~\cite{ray2021disinformation, figueira2017current, vaccari2020deepfakes}. Unethical use of this technology can lead to political instability and social violence~\cite{vaccari2020deepfakes}. On the one hand, deepfake technology continues to evolve to create more convincing and realistic fake multimedia content. Social media, on the other hand, plays a catalytic role in spreading such content. Therefore, timely detection of deepfake content is crucial to avoid any damage and loss to human society~\cite{ray2021disinformation}. 

Audiovisual deepfakes that involve multimodal manipulation are a more convincing type of forgery, with attackers attacking audio, video, or both modalities. Unimodal video forgery detectors~\cite{afchar2018mesonet,nguyen2019capsule, lutz2021deepfake, Haliassos_2021_CVPR} and spoofed audio detectors~\cite{wang2015relative, todisco2016new, patel2015combining, wu2022partially} are generally unable to identify forgeries across multiple modalities, although they may be good at detecting forgeries in the specific modality they focus on. To address this challenge, the research community has developed sophisticated tools and algorithms to detect audiovisual forgeries in videos. These specialized tools require knowledge of multimedia forensics as well as knowledge of deep learning. Furthermore, these tools do not generalize well to other unseen datasets and manipulations.

Large language models (LLMs) are a major advancement in the field of artificial intelligence. They are trained on a large amount of data and can perform well in various natural language processing (NLP) tasks such as text generation, summarization, classification, completion, sentimental analysis, machine translation, and question answering. Their applications even go beyond the aforementioned NLP tasks and can be used as writing assistants, learning tools, productivity tools, coding assistants, software development, healthcare, legal assistance, entertainment, and more. Despite being primarily designed for NLP tasks, OpenAI's ChatGPT can analyze image, audio, and video content. Taking advantage of its support for multimodal input, we studied the potential and limitations of ChatGPT for audiovisual deepfake detection.

The research questions we aimed to address in this study are as follows:
\begin{itemize}
  \item Can ChatGPT perform multimedia forensic tasks?
  \item Is ChatGPT capable of detecting forgery based on artifacts in audio and visual modalities?
  \item What is the role of prompt engineering in using ChatGPT to detect audiovisual deepfakes?
  \item Which performs better at identifying forgeries in audiovisual deepfakes, ChatGPT, humans, or AI models? 
  \item How interpretable is ChatGPT for forgery detection?
  \item What are the limitations of ChatGPT in detecting multimodal deepfakes? 
\end{itemize}

The main contributions of our work are threefold:
\begin{itemize}
  \item We explore for the first time the potential of ChatGPT for audiovisual forgery detection tasks.
  \item We compare the performance of ChatGPT with human and state-of-the-art AI models on audiovisual forgery detection tasks.
  \item We highlight the strengths and limitations of ChatGPT on audiovisual forgery detection tasks.

\end{itemize}

\section{Related Work}
The societal impact of synthetic media content has prompted research from multiple angles within the forensic science community. In~\cite{karnouskos2020artificial}, the authors investigated synthetic content from multiple perspectives such as multimedia content production, representation, media audience dynamics, gender, politics, law, and regulation, and concluded that the intersection between media and deepfake content can have multiple impacts on individuals and society. A study in~\cite{zachary2020digital} on the impact of unreliable deepfake information on voter behavior in US elections and democracy suggests multi-stakeholder partnerships and technological approaches for identifying and mitigating manipulated content on public platforms. In~\cite{carvalko2024generative}, the balance between innovation and ensuring fair protection under existing laws is explored, particularly as generative AI blurs the line between human and machine-generated works. The US FDA's regulation of AI in medical devices and the European AI Act, which classify AI applications based on potential harm, are initiatives aimed at addressing challenges and aligning AI-generated content and applications with human-centered values. This analysis is essential for developing a legal framework that addresses the ethical and practical implications of the creativity of generative AI in the legal domain. A recent study in~\cite{apolo2024beyond} investigated the risk of deepfakes in legal proceedings, where altered audiovisual evidence could compromise the integrity of justice. It highlights how deepfake technology can influence the outcome of cases based on subjective human judgment. The findings point to the need for changes to the legal framework to ensure that key judicial principles such as the presumption of innocence and the right to a fair trial are protected. Furthermore, research on human perception of audiovisual deepfake videos~\cite{hashmi2024unmasking} shows that it is difficult for people to accurately distinguish deepfake content from real videos, mainly due to the realistic visual and acoustic manipulations involved.

Audiovisual deepfakes can be broadly categorized into three types, as shown in Fig. \ref{fig:Fig1}. The first type is ``Fake Video Real Audio'' (FVRA), in which the visual frames are manipulated using techniques such as Faceswap~\cite{korshunova2017fast}, Fsgan~\cite{nirkin2019fsgan}, and Wav2lip~\cite{prajwal2020lip}, while the audio modality remains unaltered. The second type is ``Real Video Fake Audio'' (RVFA), where the video frames are authentic, but the audio modality is manipulated using techniques such as SV2TTS~\cite{jia2018transfer}, a real-time voice cloning tool that can synthesize fabricated audio content. The third type is ``Fake Video Fake Audio'' (FVFA), where both modalities are manipulated. In this type, face manipulation can be done using methods like Faceswap and Fsgan, while lip synchronization/manipulation can be achieved using Wav2lip. Additionally, cloned or synthesized audio can be integrated with visually manipulated frames to produce more realistic and convincing deepfake videos. 

The multimedia forensics community has developed several data-driven audiovisual deepfake detection methods based on multimodal feature fusion~\cite{zhou2021joint}, ensemble learning~\cite{khalid2021evaluation, hashmi2022multimodal}, and synchronization features~\cite{shahzad2022lip, shahzad2023av}. Models based on Convolutional Neural Networks (CNN), Recurrent Neural Networks (RNN), and Transformers~\cite{ilyas2023avfakenet, shahzad2023av, hashmi2023avtenet, yang2023avoid} have been widely used to detect forgeries in either modality and are trained on multimodal deepfake datasets. 
These methods provide a binary output for the input video, indicating whether the input video is genuine or spoofed. 
Disadvantages of these end-to-end learning-based methods include reliance on large datasets, heavy training, and lack of interpretability and generalization. Bias, imbalance, and lack of diversity in training data can lead to fairness, generalization, and security issues for detection models~\cite{xu2024analyzing}.

Recently, with the emergence of LLMs, the research community has begun to utilize these models to solve various tasks in different fields, beyond their original purpose. For example, while ChatGPT is primarily designed for NLP tasks, the multimodal mode in ChatGPT-4 enables it to handle multimodal inputs and analyze content from a multimodal perspective~\cite{bang2023multitask, vaikuntamultimodal, yang2023mm}. Many studies have investigated the performance of LLMs in various challenging tasks, such as image forensics~\cite{jia2024can}, facial biometrics~\cite{deandres2024good}, tampered image detection~\cite{yang2024research}, fake news detection~\cite{caramancion2023harnessing}, NLP~\cite{kocon2023chatgpt}, cheap-fake detection~\cite{wu2023cheap}, global warming~\cite{biswas2023potential}, education~\cite{he2024exploring}, public health~\cite{biswas2023role}, and medical applications~\cite{yan2023multimodal}. These studies highlight the strengths and limitations of LLMs, focusing specifically on ChatGPT's effectiveness in handling these tasks. Unlike traditional machine learning-based multimedia forensic tools, LLMs are readily accessible and can be used for multimedia forgery detection tasks. In this study, we aim to leverage the implicit knowledge embedded in ChatGPT and the generalization ability of multimodal inputs to accomplish the audiovisual deepfake detection task. 

\section{Methodology}
\begin{figure}[t]
    \centering
    \includegraphics[width=0.49\textwidth]{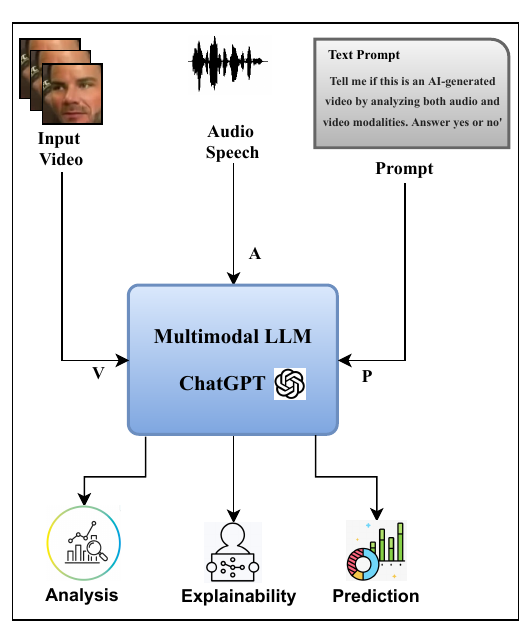}
    \caption{Illustration of the multimodal capabilities of a large language model, which takes inputs such as visual frames, audio speech, and text prompts to produce outputs that include audiovisual analysis, interpretation, and authenticity prediction.}
    \label{fig:Fig2}
\end{figure}

Fig. \ref{fig:Fig2} shows an LLM-based approach for multimodal forgery detection, where the text prompts and video with corresponding audio are used as inputs. Based on the given prompts, the model works as a black box and produces multiple analysis results on the input video, such as visual, acoustic, and audiovisual analysis and their corresponding predictions. Our goal is to evaluate the detection capabilities of ChatGPT. This model is trained on multimodal data and can be used for audiovisual forgery detection tasks. Deepfake attacks usually target high-level facial features and voice identities; therefore, we choose frontal face videos with voices to evaluate the detection performance of ChatGPT. Unlike traditional end-to-end models that leverage low-level features, LLMs provide high-level features and descriptions to analyze multimodal inputs. In this study, we used OpenAI's GPT-4 to conduct audiovisual analysis of deepfake videos. Unlike other deep learning-based models, it provides interpretability by explaining the reasoning behind the final decision, thereby increasing the transparency of the decision-making process.
\subsection{Prompt Design}
\label{sec:Prompt Design}
Inspired by previous research on LLM-based image forensics~\cite{jia2024can}, we proposed the following custom prompts, ranging from simple, binary-answer prompts to advanced, context-rich prompts designed to account for artifacts and manipulations in audiovisual content:
\begin{itemize}
  \item P1: Tell me if this is an AI-generated video by analyzing both audio and video modalities. Answer yes or no.
  \item P2: Tell me if this is a real video by analyzing both audio and video modalities. Answer yes or no.
  \item P3: Tell me the probability of this video being real by analyzing both audio and video modalities. Answer a probability score between 0 and 100.
  \item P4: Tell me the probability of this video being AI-generated by analyzing both audio and video modalities. Answer a probability score between 0 and 100.
  \item P5: Tell me if this is a real or AI-generated video by analyzing both audio and video modalities.
  \item P6: Tell me if synthesis artifacts are in the visual frames or audio. Must return with 1) yes or no only; 2) if yes, explain where the artifacts exist by answering in [region, artifacts] form.
  \item P7: I want you to work as a multimedia forensic expert for audiovisual deepfake detection. Check if the video has the artifact attribute listed in the following list, and ONLY return the attribute number in this video. The artifact list is [1-unnatural face edges; 2-inconsistent lighting and shading; 3-irregular teeth shape or texture; 4-irregular lip movement; 5-inconsistent skin texture; 6-spectral artifacts; 7-phoneme artifacts; 8-inconsistencies in speech patterns; 9-voice quality issues; 10-lack of synchronization between audio and video].
\end{itemize}

\subsection{Input of LLM Model}
We feed the videos directly into the LLM-based ChatGPT model without performing any preprocessing or transcribing the videos for analysis. The model extracts audio from the video and performs visual and acoustic analysis based on input prompts. Let
    $\text{X} = \left( \text{x}_{\text{v}}, \text{x}_{\text{a}}, \text{x}_{\text{t}} \right)$ denote the entire input, 
where
\(  \text{x}_{\text{v}} \) represents the video, \(  \text{x}_{\text{a}} \) represents the audio, and
\( \text{x}_{\text{t}} \) represents the custom text prompt.
The model outputs its final prediction as:
\begin{equation}
    \hat{y} = f_{\text{LLM}}(\text{X}) = f_{\text{LLM}}(\text{x}_{\text{v}}, \text{x}_{\text{a}}, \text{x}_{\text{t}}),
\end{equation}
where \( f_{\text{LLM}} \) is the underlying function.

\subsection{Audiovisual Analysis} Based on the given input text prompts, the model performs audio analysis by analyzing spectral features, zero crossing rate, mel-frequency cepstral coefficients (MFCC), amplitude envelope, amplitude range, median amplitude, spectral centroid, spectral rolloff, and silence ratio. For video, the model performs analysis such as blurriness, pixelation, lighting, frame difference mean, frame difference standard deviation, unnatural expression, skin texture, lip-syncing, and structural similarity index (SSIM). It also performs multimodal analysis to verify consistency between visual and audio modalities through synchronization checks. By combining unimodal acoustic and visual analyses with multimodal analysis, joint analysis is performed to reach a final prediction or suggest further manual inspection.

\subsection{Prediction Assignment}
The final prediction  \( \hat{y} \) for each input video is determined based on the following factors: yes/no response, probability score, overall conclusion, artifact-free versus artifact list, and estimated likelihood of the input video being real or fake. Fake classes are assigned label 1, while label 0 represents real classes.

\subsection{ChatGPT vs Human vs AI Models} To understand the detection capabilities of ChatGPT, humans, and AI models, we follow the study in~\cite{hashmi2024unmasking}, where the authors reported a comparative analysis between humans and deep-learning-based multimodal forensic models. 
Their results concluded that state-of-the-art AI models surpass humans in detecting multimodal deepfakes. Furthermore, participants often showed overconfidence in their detections, with their average accuracy being lower than their confidence level. 
To evaluate the detection performance of ChatGPT, we selected the same video subset used in~\cite{hashmi2024unmasking} from the FakeAVCeleb dataset~\cite{khalid2021fakeavceleb} for a fair comparison.

\section{Experiments and Results}
\subsection{Dataset Selection}
Following the study in~\cite{hashmi2024unmasking}, we selected the same 40 videos from the FakeAVCeleb dataset~\cite{khalid2021fakeavceleb} to allow for a fair comparison with humans and state-of-the-art multimodal forensic models. The 40 videos contain 20 real videos and 20 fake videos, representing each class equally. To eliminate gender bias, each class contains an equal number of male and female videos. 

\subsection{Evaluation Metrics}
\label{sec: Evaluation Metrics}
For evaluation, we calculate precision, recall, F1-score, and accuracy, which are defined as, 
\begin{equation}\label{eq_1}
    \text{Precision} = \frac{TP}{TP+FP},
\end{equation}

\begin{equation}\label{eq_2}
    \text{Recall} = \frac{TP}{TP+FN},
\end{equation}

\begin{equation}\label{eq_3}
    \text{F1} = \frac{2 \times \text{Precision} \times \text{Recall}}{\text{Precision} + \text{Recall}},
\end{equation}

\begin{equation}\label{eq_4}
    \text{Accuracy} = \frac{TP + TN}{TP + TN + FP + FN}.
\end{equation}
where ${TP}$, ${TN}$, ${FP}$, and ${FN}$ stand for True Positive (fake videos correctly detected as fake), True Negative (real videos correctly detected as real), False Positive (real videos incorrectly detected as fake), and False Negative (fake videos incorrectly detected as real), respectively.
Additionally, we calculate the rejection rate to evaluate the effectiveness of input text prompts: 
\begin{equation}\label{eq_5}
\text{Rejection Rate}= {\frac{\text{Number of Rejected Prompts}}{\text{Total Number of Prompts}} \times 100}.
\end{equation}

\begin{table}[t]
\caption{Precision, Recall, F1 Score, Rejection Rate, and Accuracy for different text prompts (P1-P7).}
\centering
\begin{tabular}{|l|c|c|c|c|c|c}
\hline
Prompt & Precision & Recall & F1 Score & Rejection Rate & Accuracy \\ \hline
P1    & 0.583    & 0.368  & 0.451     & 7.50           & 54.0    \\ \hline
P2    & 0.625    & 0.250  & 0.357     & 2.50           & 53.8    \\ \hline
P3    & 0.625    & 0.250  & 0.357     & 0.00           & 55.0    \\ \hline
P4    & 0.571    & 0.600  & 0.585     & 0.00           & 57.5    \\ \hline
P5    & 0.571    & 0.200  & 0.300     & 0.00           & 52.5    \\ \hline
P6    & 0.607    & 0.850  & 0.706     & 0.00           & 65.0    \\ \hline
P7    & 0.600    & 0.900  & 0.720     & 0.00           & 65.0    \\ \hline
\end{tabular}
\vspace{0.1cm} 
\label{table:table_1}
\vspace{-4mm}
\end{table}

\subsection{Results} 
\subsubsection{Comparing Different Text Prompts} Table \ref{table:table_1} lists the performance of different text prompts (P1-P7 in Section~\ref{sec:Prompt Design}). Fig. \ref{fig:Fig3} shows a bar graph comparison of ${TP}$, ${TN}$, ${FP}$, and ${FN}$ (see Section~\ref{sec: Evaluation Metrics}) for different text prompts. Additionally, Fig. \ref{fig:P1} shows an example of custom text prompts, ChatGPT audiovisual analysis, corresponding predication, and ground truth label. Next, we analyze each text prompt in detail and discuss the results.
\begin{itemize}
\begin{figure}[t]
\centering
\includegraphics[width=0.5\textwidth]{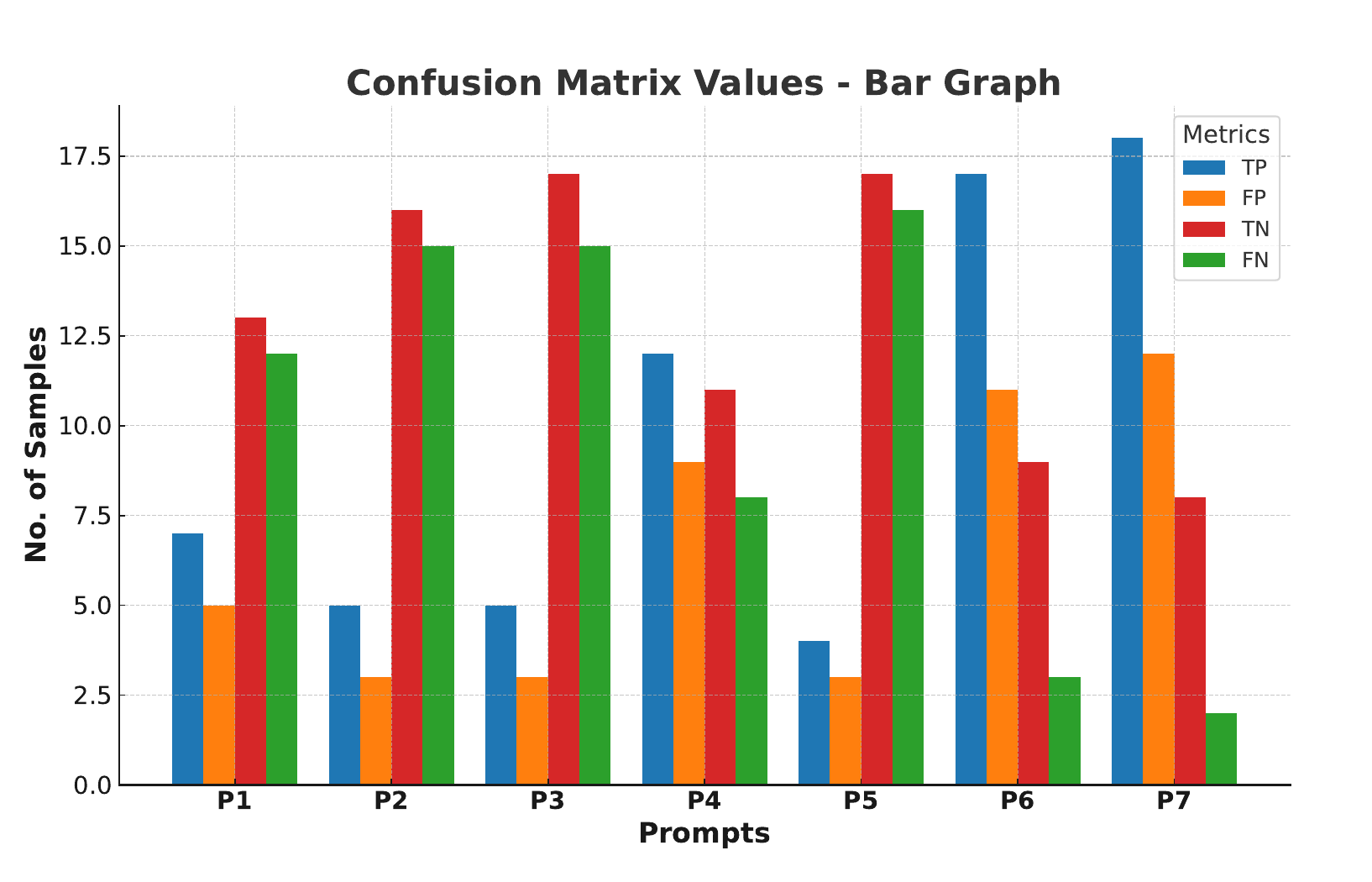}
\caption{Bar graph comparing the number of True Positives, False Positives, True Negatives, and False Negatives for different text prompts.}
\label{fig:Fig3}
\end{figure}

\begin{figure*}[t]
\centering
    \fbox{\includegraphics[width=0.97\textwidth]{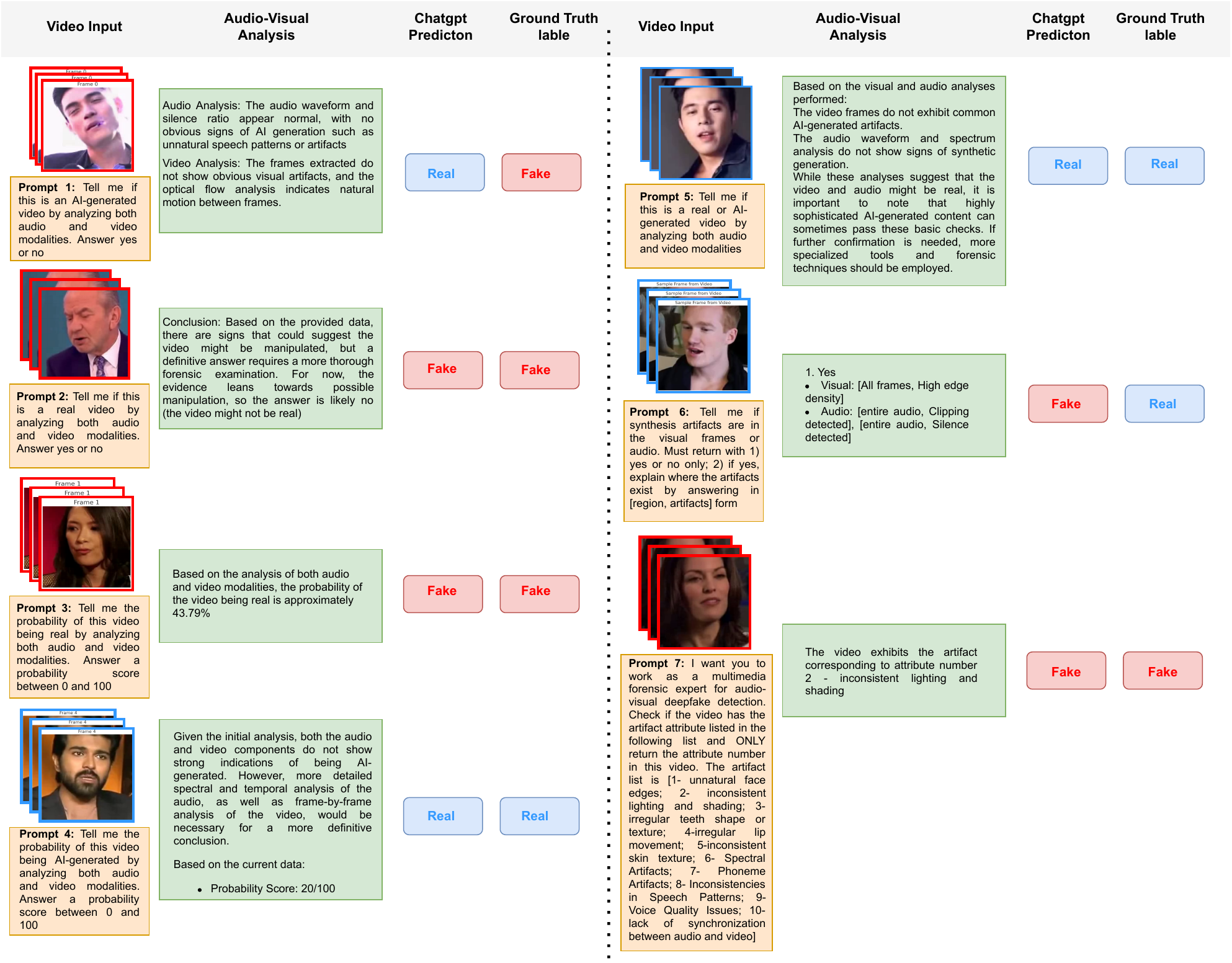}}
    \caption{Demonstration of ChatGPT responses, which takes video and text prompts as input and produces audiovisual analysis, including explanations and authenticity predictions.}
    \label{fig:P1}
\vspace{-2mm}
\end{figure*}

\item  Prompt P1: As can be seen from Table \ref{table:table_1}, the accuracy of P1 reaches 54.0\%, which is only slightly higher than the 50\% of random guessing, indicating that it is less effective in guiding the model to make accurate predictions. Its simplicity and lack of necessary information resulted in a rejection rate as high as 7.50\%, preventing the model from making predictions for every video input. The numbers of $TP$, $FP$, $TN$, and $FN$ are 7, 5, 13, and 12 respectively. In the deepfake video detection task, a higher $TP$ value is desirable. However, instead of obtaining higher $TP$, P1 produced more $TN$, resulting in a lower recall of 0.368. 

\item Prompt P2: Similar to P1, P2 generates a binary response as to whether the video is real or not. Its lack of contextual information resulted in an even lower accuracy of 53.8\%, but it was better than P1 in terms of rejection rate, which was 2.50\%. The numbers of $TP$, $FP$, $TN$, and $FN$ are 5, 3, 16, and 15, respectively. In terms of $TP$, P2 performed worse than P1. Due to the lower number of $FP$, its precision was slightly better than P1 at 0.625, while P2 had a higher number of $FN$, resulting in a lower recall than P1 at 0.250.      

\item Prompt P3: Unlike the binary response output in P1 and P2, P3 requires the model to return a probabilistic output, which results in an accuracy of 55.0\% and a rejection rate of 0\%, slightly better performance than P1 and P2. The numbers of $TP$, $FP$, $TN$, and $FN$ are 5, 3, 17, and 15, respectively. 
Compared to P2, $TN$ increases by 1.
The precision and recall rates are 0.625 and 0.250, the same as P2. 

\item Prompt P4: P4 has improved accuracy compared to the previous three prompts. This prompt contains the term ``AI-generated'' and requires a probability score, resulting in better performance and increased accuracy to 57.5\%. The numbers of $TP$, $FP$, $TN$, and $FN$ are 12, 9, 11, and 8, respectively. The precision rate is 0.571, which is slightly lower due to the higher number of $FP$ compared to the previous three prompts. However, due to the lower number of $FN$, P4 achieves a better recall rate of 0.600. 

\item Prompt P5: Like P1, P2, and P3, P5 appeared to be less effective due to its lack of specificity and manipulation details. P5 achieved an accuracy of 52.5\%. The numbers of $TP$, $FP$, $TN$, and $FN$ are 4, 3, 17, and 16, respectively. P5 has the same precision as P4, but its recall is poor at 0.200. The main reason for the extremely low recall rate is the small number of $TP$, only 4 out of 20 fake samples were correctly detected as fakes.

\item Prompt P6: With an accuracy of 65.0\%, higher precision and recall, and a rejection rate of 0\%, P6 is by far the best performing prompt. Unlike the previous prompts, P6 focuses on visual and acoustic artifacts present in visual and audio modalities, allowing the underlying multimodal model to make the final prediction/decision more effectively. The numbers of $TP$, $FP$, $TN$, and $FN$ are 17, 11, 9, and 3, respectively. P6 has a higher number of $TP$ and $TN$ than the previous prompts.  

\item Prompt P7: Compared to P6, the contextual details in P7 text prompt narrow the focus of the LLM model to specific artifacts and manipulations in both modalities, yielding more accurate and reliable results. P7 achieved an accuracy of 65.0\%, a rejection rate of 0\%, and the highest recall among all text prompts due to the larger number of $TP$. The numbers of $TP$, $FP$, $TN$, and $FN$ are 18, 12, 8, and 2, respectively.
\end{itemize}

In summary, prompts based on simple binary responses often lack the necessary clarity and details to effectively leverage the multimodal knowledge of the LLM. Therefore, prompts like P1 to P5 resulted in lower accuracy. In contrast, more context-rich, artifact-based, and detailed-oriented prompts, such as P6 and P7, outperformed other simpler prompts. These more effective prompts leverage the multimodal capabilities and underlying knowledge of the LLM, yielding detection results aware of specific artifacts and manipulations in both modalities. 

\subsubsection{Comparing ChatGPT with Human and AI models}
The detection performance of human evaluators and various state-of-the-art deep learning-based models was compared in~\cite{hashmi2024unmasking}, as summarized in Table~\ref{table:accuracy_rejection}. 
In the human subjective test, each subject evaluated the same set of videos twice, in a different playback order. Phase 1 in the table represents the average accuracy of all testers in the first round, and Phase 2 represents the average accuracy of the second round. The results showed that humans performed better in the second round, but the difference between the two rounds was not significant.
The average accuracy of human evaluators was 65.64\%, which serves as the baseline for comparison in our study. Among AI models, Lipforensics~\cite{Haliassos_2021_CVPR} focuses on semantic inconsistencies in the mouth region and shows strong performance with an accuracy of 92.50\%. The AV-Lip-Sync model~\cite{shahzad2022lip} exploits synchronization between audiovisual modalities and achieves an accuracy of 87.50\%, which is slightly lower than the accuracy of Lipforensics. The remaining three models AV-Lip-Sync+~\cite{shahzad2023av}, CNN-Ensemble~\cite{hashmi2022multimodal}, and AVTENet~\cite{hashmi2023avtenet} all achieved a higher accuracy of 97.50\%. 
The comparison results in Table~\ref{table:accuracy_rejection} show that ChatGP performs on par with humans when provided with appropriate prompts, but both perform much worse than today's AI detection models.
The higher accuracy of these deepfake detection models is attributed to their training on the multimodal FakeAVCeleb dataset.

\begin{table}[t]
\vspace{-4mm}
\centering
\caption{Comparison of detection accuracy between humans, AI models, and ChatGPT.}
\begin{tabular}{|c|c|c|c|}
\hline
\textbf{Category} & \textbf{Method} & \textbf{Accuracy (\%)} & \textbf{Rejection} \\
                                                                    &&&\textbf{Rate (\%)} \\
\hline
\multirow{3}{*}{Human} & Human (Phase I)~\cite{hashmi2024unmasking} & 63.30 & - \\
 & Human (Phase II)~\cite{hashmi2024unmasking} & 67.98 & - \\
 & Human (Overall)~\cite{hashmi2024unmasking} & 65.64 & - \\
\hline
\multirow{5}{*}{AI Models} & LipForensics~\cite{Haliassos_2021_CVPR} & 92.50 & - \\
 & AV-Lip-Sync~\cite{shahzad2022lip}& 87.50 & - \\
 & AV-Lip-Sync+~\cite{shahzad2023av}& 97.50 & - \\
 & CNN-Ensemble~\cite{hashmi2022multimodal}& 97.50 & - \\
 & AVTENet~\cite{hashmi2023avtenet}& 97.50 & - \\
\hline
\multirow{7}{*}{ChatGPT} & P1 & 54.05 & 7.50 \\
 & P2 & 53.85 & 2.50 \\
 & P3 & 55.00 & 0.00 \\
 & P4 & 57.50 & 0.00 \\
 & P5 & 52.50 & 0.00 \\
 & P6 & 65.00 & 0.00 \\
 & P7 &  65.00 & 0.00 \\
\hline
\end{tabular}
\vspace{0.3cm} 
\label{table:accuracy_rejection}
\vspace{-6mm}
\end{table}

\begin{table}[b]
\vspace{-6mm}
\caption{Precision, Recall, F1 Score, Rejection Rate, and Accuracy for different video-only mention prompts (P1-P7).}
\centering
\begin{tabular}{|l|c|c|c|c|c|c}
\hline
Prompt & Precision & Recall & F1 Score & Rejection Rate & Accuracy \\ \hline
P1    & 0.545    & 0.500  & 0.522     & 52.5           & 42.11    \\ \hline
P2    & 0.467    & 0.500  & 0.483     & 40.0          & 37.50    \\ \hline
P3    & 0.542    & 0.813  & 0.650     & 27.5           & 51.72   \\ \hline
P4    & 0.480    & 0.750  & 0.585     & 17.5           & 48.48    \\ \hline
P5    & 0.545    & 0.800  & 0.649     & 22.5           & 56.66    \\ \hline
P6    & 0.600    & 0.474  & 0.529     & 2.50           & 57.89    \\ \hline
P7    & 0.333    & 0.167  & 0.222     & 10.0           & 41.67    \\ \hline
\end{tabular}
\vspace{0.1cm} 
\label{table:ablation_accuracy_rejection}
\end{table}

\section{Ablation study}
\subsection{Effectiveness of Prompts}
Initially, we tested some basic prompts by mentioning only ``video'' and no mention of ``audio'', and executed custom prompts sequentially within one session. The following are the custom text prompts:
\begin{itemize}
    \item Tell me if this is an AI-generated video. Answer yes or no.
\item Tell me if this is a real video. Answer yes or no.
\item Tell me the probability of this video being real. Answer a probability score between 0 and 100.
\item Tell me the probability of this video being AI-generated. Answer a probability score between 0 and 100.
\item Tell me if this is a real or AI-generated video.
\item Tell me if synthesis artifacts are in the face. Must return with 1) yes or no only; 2) if yes, explain where the artifacts exist by answering in [region, artifacts] form.
\item I want you to work as a video forensic expert for AI-generated faces. Check if the video has the artifact attribute in the following list and ONLY return the attribute number in this image. The artifact list is [1-asymmetric eye iris; 2-irregular reflection; 3-irregular teeth shape or texture; 4-irregular ears or earrings; 5-strange hair texture; 6-inconsistent skin texture; 7-inconsistent lighting and shading; 8-strange background; 9-unnatural edges; 10-lack of synchronization between audio and video].
\end{itemize}

As can be seen from~Table \ref{table:ablation_accuracy_rejection}, the rejection rate is higher and the accuracy is lower, compared with the results in Table \ref{table:table_1}. Based on these results, we made several observations. First, only mentioning the video in the prompt causes the model to focus mainly on visual frames without analyzing the audio modality. To obtain the desirable output from an LLM-based model, prompts need to be specific and context-rich. 
Second, when prompts are fed sequentially, the model takes into account the context of the results of previous prompts, affecting its response to the current prompt. To obtain independent and unbiased results from prompts, we must feed the input video and prompt independently within a session to eliminate the effects of contextual bias from previous prompts.
Third, our experiments show that prompts must contain terms relevant to acoustic analysis in order for the model to effectively analyze the audiovisual content in a given video.



\subsection{Failure Case Study}
The reasons for detection failure vary depending on both the input prompt and video content. Through our experiments and careful analysis, we observed several factors that lead to inaccurate decisions in the multimodal ChatGPT model. 

One factor is a high silence ratio in the speech content, which may indicate robotic/synthetic audio since the speech generation pipeline excludes environmental noise. However, videos with clean/enhanced speech do not always indicate synthesis or voice spoofing. Conversely, adding synthetic environmental noise to clean audio can mislead the model, leading to inaccurate predictions. The high silence rate combined with unnatural pauses in the acoustic modality can lead to an increased number of false positives in the model.

While unimodal/multimodal deep features and audiovisual correlation features are effective in various multimodal tasks, ChatGPT mainly relies on hand-crafted features and traditional functions in computer vision and speech processing libraries, including OpenCV,  librosa, numpy, wav, and skimage, for visual and acoustic analysis. Furthermore, existing deep learning-based pretrained foundation models~\cite{shi2021learning, dosovitskiy2020image, Arnab_2021_ICCV, gong21b_interspeech} and frameworks (such as Tensorflow or Pytorch) are not used to analyze video content for possible artifacts and forgeries. These two shortcomings limit the ChatGPT method from effectively analyzing video content and result in poor performance compared to state-of-the-art forensic models. 

In the context of audiovisual video forgery detection, if any modality (audio or video) is fake, the final prediction should be classified as fake. However, we observed that the overall probability score, calculated as the average of the audio and video scores, can lead the model to make incorrect predictions. If the score of one of the modalities dominates, the final prediction tends to reflect that modality, compromising the overall accuracy and classification results. 

\section{Limitations} Although LLM-based models are superior to end-end learning-based black box models in terms of generalization, interpretability, and intuitive user interface for end users, they still have limitations. LLM-based models require domain knowledge for multimedia forensics tasks to design more effective prompts to exploit their underlying multimodal capabilities. Simple binary prompts are ineffective and yield lower accuracy and higher rejection rates. Furthermore, ChatGPT uses traditional techniques to analyze forgery in audiovisual content and has no access to pretrained models specifically trained for multimodal forgery detection tasks, resulting in lower accuracy even when the text prompts are effective and contextually rich. Given these limitations, the multimedia forensics community must focus on cutting-edge, end-to-end learning-based techniques to develop more robust, generalizable, and explainable audiovisual deepfake detectors.

\section{Conclusions}
In this study, we investigated the detection capabilities of a large language model (LLM) (i.e., ChatGPT) in the multimodal forgery detection task. We compared its performance with that of end-to-end multimedia forensic methods and human capabilities. Our results showed that, although ChatGPT was not explicitly designed for multimedia forgery detection tasks, its performance was comparable to human detection performance, demonstrating its potential in this field. A notable advantage of using LLMs in video forensics is their ability to generalize effectively because these models are learned from a wide range of datasets, unlike end-to-end models that are typically learned from specific video deepfake datasets. Additionally, LLMs provide superior interpretability compared to deep learning-based forensic methods, which, while capable of identifying specific visual and acoustic artifacts, typically serve as black-box models with limited interpretability. In future work, we aim to combine LLM-based models with deep learning-based forensic models to enhance interpretability and further contribute more interpretable and transparent deepfake detection tools to the forensics community.



\ifCLASSOPTIONcaptionsoff
  \newpage
\fi
\bibliographystyle{elsarticle-num}
\bibliography{references}
\end{document}